\documentclass[sigconf]{acmart}

\usepackage{balance}
\usepackage{float}
\usepackage{scalerel}
\usepackage{amsfonts}       
\usepackage{graphicx}
\usepackage{comment}
\usepackage{caption}
\usepackage{subcaption}
\usepackage{amsmath}
\usepackage{amssymb}
\usepackage[linesnumbered,ruled,vlined]{algorithm2e}

\acmSubmissionID{aiespp46}

\copyrightyear{2020}
\acmYear{2020}
\setcopyright{acmlicensed}
\acmConference[AIES '20]{Proceedings of the 2020 AAAI/ACM Conference on AI, Ethics, and Society}{February 7--8, 2020}{New York, NY, USA}
\acmBooktitle{Proceedings of the 2020 AAAI/ACM Conference on AI, Ethics, and Society (AIES '20), February 7--8, 2020, New York, NY, USA}
\acmPrice{15.00}
\acmDOI{10.1145/3375627.3375850}
\acmISBN{978-1-4503-7110-0/20/02}

\begin{document}

\fancyhead{}

\title{FACE:\\Feasible and Actionable Counterfactual Explanations}

\author{Rafael Poyiadzi}
\affiliation{%
  \institution{University of Bristol}
  \city{Bristol}
  \country{United Kingdom}}
\email{rp13102@bristol.ac.uk}

\author{Kacper Sokol}
\affiliation{%
  \institution{University of Bristol}
  \city{Bristol}
  \country{United Kingdom}}
\email{K.Sokol@bristol.ac.uk}

\author{Raul Santos-Rodriguez}
\affiliation{%
  \institution{University of Bristol}
  \city{Bristol}
  \country{United Kingdom}}
\email{enrsr@bristol.ac.uk}

\author{Tijl De Bie}
\affiliation{%
  \institution{University of Ghent}
  \city{Ghent}
  \country{Belgium}}
\email{tijl.debie@ugent.be}

\author{Peter Flach}
\affiliation{%
  \institution{University of Bristol}
  \city{Bristol}
  \country{United Kingdom}}
\email{Peter.Flach@bristol.ac.uk}

%
\renewcommand{\shortauthors}{Poyiadzi and Sokol, et al.}

%
\begin{abstract}
Work in Counterfactual Explanations tends to focus on the principle of ``the closest possible world'' that identifies small changes leading to the desired outcome. In this paper we argue that while this approach might initially seem intuitively appealing it exhibits shortcomings not addressed in the current literature. First, a counterfactual example generated by the state-of-the-art systems is not necessarily representative of the underlying data distribution, and may therefore prescribe unachievable goals (e.g., an unsuccessful life insurance applicant with severe disability may be advised to do more sports). Secondly, the counterfactuals may not be based on a ``feasible path'' between the current state of the subject and the suggested one, making actionable recourse infeasible (e.g., low-skilled unsuccessful mortgage applicants may be told to double their salary, which may be hard without first increasing their skill level). These two shortcomings may render counterfactual explanations impractical and sometimes outright offensive.
To address these two major flaws, first of all, we propose a new line of Counterfactual Explanations research aimed at providing actionable and feasible paths to transform a selected instance into one that meets a certain goal.
Secondly, we propose \textbf{FACE}: an algorithmically sound way of uncovering these ``feasible paths'' based on the shortest path distances defined via density-weighted metrics. Our approach generates counterfactuals that are coherent with the underlying data distribution and supported by the ``feasible paths'' of change, which are achievable and can be tailored to the problem at hand.
\end{abstract}

%
%




%

%

%
\maketitle

\section{Introduction}
In this paper we are concerned with Counterfactual and Contrastive Explanations (CE) \cite{van2018contrastive} that fall under the category of \textit{Example-Based Reasoning}. While other approaches in the field of Machine Learning Interpretability \cite{ribeiro2016should,lundberg2017unified,sokol2019blimey} aim at answering: ``Why has my loan been declined?'', CE aim at answering: ``What do I need to do for my loan to be accepted?'' 

\citet{wachter2017counterfactual} propose three aims of explanations with respect to their audience:
\begin{enumerate}
    \item to inform and help the explainee understand why a particular decision was reached,
    \item to provide grounds to contest adverse decisions, and
    \item to understand what could be changed to receive a desired result in the future, based on the current decision-making model.
\end{enumerate}
\begin{figure}[t!]
    \centering
    \includegraphics[width=1.\linewidth]{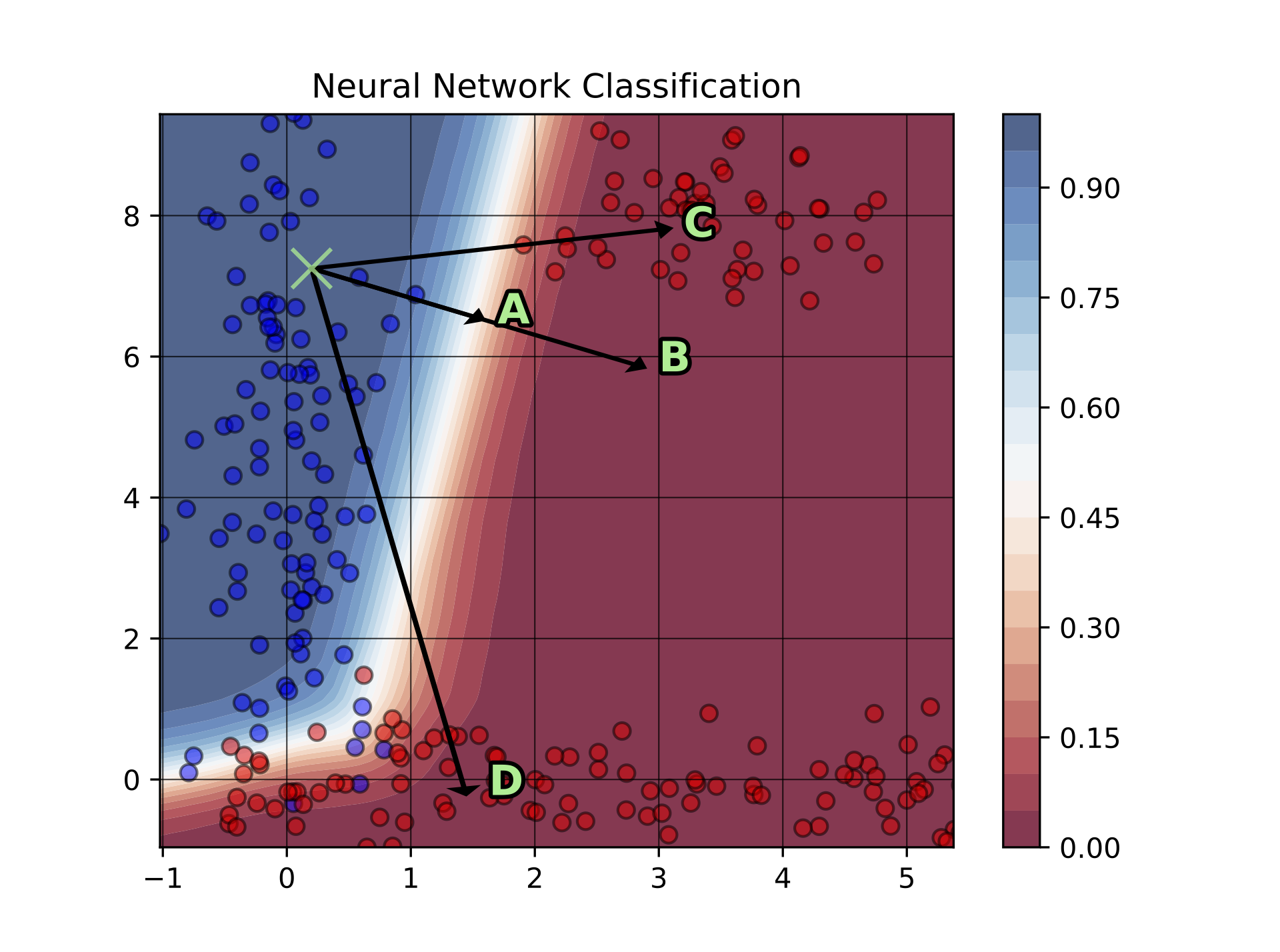}
    \caption{$A$, $B$, $C$ and $D$ are four viable counterfactuals of $\times$, all satisfying the condition of having a different predicted class. We argue that $D$ is the best choice. $A$ is the result of minimising the $l_2$-norm and $B$ is a generic data point that has a large classification margin. Nevertheless, both $A$ and $B$ lie in a \emph{low density region}. $C$ and $D$ do not share the shortcomings of $A$ and $B$: they lie in high-density regions and have a relatively large classification margins. The major difference between $C$ and $D$ is the connection between $\times$ and $D$ via a high-density path, indicating that it is feasible for the original instance to be transformed into $D$ despite $C$ being simply closer.}
    \label{fig:problems}
\end{figure}
Counterfactual explanations achieve all three of these aims \cite{wachter2017counterfactual}.
However, a na\"ive application of the last one -- the principle of ``the closest possible world'' that prescribes small changes that lead to the desired outcome -- may yield inadequate results.
Firstly, a counterfactual generated by a state-of-the-art explainability system is not necessarily representative of the underlying data distribution and may prescribe unachievable goals. This shortcoming is illustrated in Figure~\ref{fig:problems}, where points $A$ and $B$ -- both close to the explained data point $\times$ with respect to the $l_2$-norm -- achieve the desired prediction, however they lie in a low-density region. This last observation undermines
the practical feasibility of $A$ and $B$ since there are no precedents of similar instances in the data.
Secondly, counterfactuals provided by current approaches may not allow for a \textit{feasible path} between the initial instance and the suggested counterfactual making actionable recourse impractical. This argument is illustrated with point $D$ in Figure~\ref{fig:problems}, which we argue is a more actionable counterfactual than $C$.
Both these discoveries have prompted us to establish a new
line of research for Counterfactual Explanations: providing \emph{actionable} and \emph{feasible} paths to transform a certain data point into one that meets certain goals (e.g., belong to a desirable class).

The contribution of this paper is twofold. We first critique the existing line of research on Counterfactual Explanations by pointing out the shortcomings of dismissing the nature of the target counterfactual and its (real-life) context. We point out that existing research is not aligned with real-world applications (e.g., offering a \emph{useful} counterfactual advice to customers who have been denied loans). To overcome this challenge we identify two essential properties of counterfactual explanations: \emph{feasibility} and \emph{actionability}, which motivate a new line of research concerned with providing high-density paths of change.
Secondly, we propose a novel, well-founded approach to generating feasible and actionable counterfactual explanations that respect the underlying data distribution and are connected via high-density paths (based on the shortest path distances defined via density-weighted metrics) to the explained instance.
Our approach -- which we call \emph{Feasible and Actionable Counterfactual Explanations} (\textbf{FACE}) -- mitigates all of the risks associated with the explanations produced by the current line of research.

We support our claims by discussing how overlooking these premises could lead to ``unachievable goals'' with undesired consequences such as a loss of the end user's trust. Furthermore, we show that our algorithmic contribution to generating feasible and actionable counterfactuals is non-trivial as the resulting explanations come from dense regions and are connected with high-density paths to the original instance. Therefore, the explanations are coherent with the underlying data distribution and can be tailored to the user by customising the ``feasible paths'' of change. In Section~\ref{counterfactuals} we establish the links and differences of our method with similar approaches in the literature. Section~\ref{methods} introduces our methodology and Section~\ref{experiments} presents our experimental results. %
In Section~\ref{related_work} we discuss related work and in Section~\ref{discussion} we conclude our paper with a discussion.

\section{Counterfactual Explanations\label{counterfactuals}}

In this section we motivate the need for a new approach (given the current literature) that ensures usefulness of counterfactual explanations in practice.
Firstly, the nature of the target instance -- the derived counterfactual example -- is not taken into account. This may lead to a case where the target instance is not representative of the underlying data distribution, for example, it is located in a low density region, and thus can be considered an outlier. In addition to being poor explanations, such counterfactuals are at risk of
harming the explainee
by suggesting a change of which the future outcome is highly uncertain, as classifiers tend to be less reliable in sparsely populated regions of the data space, especially close to a decision boundary.
Points $A$ and $B$ shown in Figure~\ref{fig:problems} are examples of this major drawback. The uncertainty in a prediction, coming either from a low classification margin or due to low density of a region, should be of utmost importance when generating a counterfactual.

Beyond feasibility and actionability, it is also important to examine the model's confidence of predictions as it may contribute to issues with a delayed impact \cite{liu2018delayed}. For example, consider a person who had his loan application rejected and wants to know what changes to make for his application to be accepted next time. If this person is handed a counterfactual explanation and implements the proposed changes, his loan application will be accepted. However, if the \textit{new state} of the subject (the proposed counterfactual) is in a region of high uncertainty, then there exists a high risk that this individual will default. 


Furthermore, the desiderata presented by \citet{wachter2017counterfactual} do not account for the extent to which the change -- a transformation from the current state to the suggested counterfactual state -- is feasible. %
``Counterfactual thinking'' refers to the concept of hypothesising what would have happened had something been done differently \cite{contrastive_thesis}, i.e., ``Had I done $X$ instead of $Y$, would the outcome still be $Z$?'' %
However, when adapting this concept to Machine Learning applications, e.g., see \citet{contrastive_thesis}, the outcome is usually decided prior to finding a counterfactual cause. What has been overlooked by the Interpretable Machine Learning community is that the aim of a counterfactual explanation is for the explainee to \textit{actually try and make the change} given the actionable nature of the explanation. A customer whose loan application has been rejected would (probably) disregard a counterfactual explanation conditioned on him being 10 years younger.


The current state-of-the-art solutions do not satisfy the three requirements proposed by \citet{wachter2017counterfactual}, which we believe are critical for actionability and thus practical utility of counterfactual explanations.
To remedy this situation we propose to following objectives for counterfactual explanations in addition to the inherent requirement of these instances belonging to the desired class:
\begin{enumerate}
    \item feasibility of the counterfactual data point,
    \item continuity and feasibility of the path linking it with the data point being explained, and
    \item high density along this path and its relatively short length.
\end{enumerate}

\section{Feasible Counterfactuals\label{methods}}
Before presenting \textbf{FACE} we introduce the necessary notation and background for completeness (see \citet{alamgir2012shortest} and references therein for an in-depth presentation of this topic). We then show how different variants of our approach affect its performance and the quality of generated counterfactuals.

\subsection{Background}
Let $\mathcal{X} \subseteq \mathbb{R}^d$ denote the input space and let $\{\boldsymbol{x}_i\}_{i=1}^{N} \in \mathcal{X}$ be an independent and identically distributed  sample from a density $p$. Also, let $f$ be a positive scalar function defined on $\mathcal{X}$ and let $\gamma$ denote a path connecting $\boldsymbol{x}_i$ to $\boldsymbol{x}_j$, then the $f$-length of the path is denoted by the \textit{line integral} along $\gamma$ with respect to $f$:\footnote{We assume that $\mathcal{X}$ is endowed with a density function $p$ with respect to the Lebesgue measure, where $p$ is $L$-Lipschitz continuous with $L > 0$.}
\begin{equation}
    \mathcal{D}_{f, \gamma} = \int_{\gamma} f(\gamma(t)) \cdot |\gamma'(t)| dt\text{.}
\end{equation}
The path with the minimum $f$-length is called the $f$-geodesic, and its $f$-length is denoted by $\mathcal{D}_{f, \gamma^\star}$.

Consider a geometric graph $G = (V, E, W)$ with vertices $V$, edges $E$ and (edge) weights $W$. The vertices correspond to the sampled instances (training data) and edges connect the ones that are close with respect to a chosen metric, which value (a measure of closeness) is encoded in the (edge) weights. We use the notation $i \sim j$ to indicate a presence of an edge connecting $\boldsymbol{x}_i$ and $\boldsymbol{x}_j$, with the corresponding weight $w_{ij}$; and $i \nsim j$ to mark that $\boldsymbol{x}_i$ and $\boldsymbol{x}_j$ are not directly connected, in which case the weight is assumed to be $w_{ij} = 0$.

Let $f$ depend on $\boldsymbol{x}$ through the density $p$ with $f_p(x) := \tilde{f}(p(x))$. Then, the $f$-length of a curve $\gamma: \left[\alpha, \beta\right] \rightarrow \mathcal{X}$ can be approximated by a Riemann sum of a partition of $\left[\alpha, \beta\right]$ in sub-intervals $[t_{i-1}, t_{i}]$ (with $t_0=\alpha$ and $t_N=\beta$):
\begin{equation*}
    \hat{\mathcal{D}}_{f, \gamma} = \sum_{i=1}^{N} f_p\Big(\frac{\gamma(t_{i-1}) + \gamma(t_{i})}{2}\Big) \cdot \|\gamma(t_{i-1}) - \gamma(t_{i})\| \text{.}
\end{equation*}
As the partition becomes finer, $\hat{\mathcal{D}}_{f, \gamma}$ converges to $\mathcal{D}_{f, \gamma}$ \cite[Chapter 3]{gamelin2003complex}. This suggests using weights of the form:
\begin{align*}
    &w_{ij} = f_p\Big(\frac{\boldsymbol{x}_i + \boldsymbol{x}_j}{2}\Big)\cdot \|\boldsymbol{x}_i - \boldsymbol{x}_j\|\text{,}\\
    &\textrm{when} \hspace{5mm} \|\boldsymbol{x}_i - \boldsymbol{x}_j\| \leq \epsilon \text{.}
\end{align*}
In the case that the condition does not hold, $w_{ij} = 0$. The true density $p$ is rarely known but \citet{orlitsky2005estimating} show that using a \textit{Kernel Density Estimator} (KDE) $\hat{p}$ instead will converge to the $f$-distance. \citet{orlitsky2005estimating} also show how to assign weights to edges while avoiding the need to perform density estimation altogether. Their results apply to two graph constructions, namely, a $k$-NN graph and an $\epsilon$-graph. In summary, for the three approaches the weights can be assigned as follows:
\begin{align}
    &w_{ij} = f_{\hat{p}}\Big(\frac{\boldsymbol{x}_i + \boldsymbol{x}_j}{2}\Big)\cdot \|\boldsymbol{x}_i - \boldsymbol{x}_j\| \hspace{25mm} \textrm{($KDE$)} \label{eq_kde}\\
    &w_{ij} = \tilde{f}\Big(\frac{r}{\|\boldsymbol{x}_i - \boldsymbol{x}_j\|}\Big) \cdot \|\boldsymbol{x}_i - \boldsymbol{x}_j\|, \hspace{3mm} r = \frac{k}{N \cdot \eta_d} \hspace{3mm} \textrm{($k$-NN)} \label{eq_knn}\\[2pt]
    &w_{ij} = \tilde{f}\Big(\frac{\epsilon^d}{\|\boldsymbol{x}_i - \boldsymbol{x}_j\|}\Big) \cdot \|\boldsymbol{x}_i - \boldsymbol{x}_j\| \hspace{20mm} \textrm{($\epsilon$-graph)} \label{eq_egraph}\\[3pt]
    &\hspace{10mm}\textrm{when} \hspace{5mm} \|\boldsymbol{x}_i - \boldsymbol{x}_j\| \leq \epsilon \nonumber
\end{align}
where $\eta_d$ denotes the volume of a sphere with a unit radius in $\mathbb{R}^d$. Similarly to above, if the condition does not hold, $w_{ij} = 0$.

\subsection{The FACE Algorithm}

Building up on this background we introduce the \textbf{FACE} algorithm. It uses $f$-distance to quantify the trade-off between the path length and the density along this path, which can subsequently be minimised using a shortest path algorithm by approximating the $f$-distance by means of a finite graph over the data set.
Moreover, \textbf{FACE} allows the user to impose additional feasibility and classifier confidence constraints in a natural and intuitive manner. 

Firstly, a graph over the data points is constructed based on one of the three approaches: $KDE$, $k$-NN or $\epsilon$-graph. The user then decides on the properties of the target instance (i.e., the counterfactual): the prediction threshold -- a lower bound on prediction confidence outputted by the model, and the density (or its proxy) threshold. This part of the algorithm is described in Algorithm~\ref{algo1}, which assumes access to a $KDE$.

To generate a counterfactual, \textbf{FACE} must be given its expected class. Optionally, the counterfactual can be additionally constrained by means of: a subjective prediction confidence threshold ($t_p$), a density threshold ($t_d$), a custom weight function ($w$), and a custom conditions function ($c$), which determines if a transition from a data point to its neighbour is feasible.\footnote{Domain knowledge of this form (e.g., immutable features such as sex or conditionally immutable changes such as age, which are only allowed to change in one direction) are incorporated within the \emph{conditions function} $c(\cdot, \cdot)$. This knowledge is \emph{essential} if the desired counterfactual is to be useful.} Subject to the new weight function and conditions function, if possible, the graph is updated by removing appropriate edges; otherwise a new graph is constructed.\footnote{If the explainee wants to provide a custom cost function for the feature value changes, e.g., the cost of changing a job is twice that of change a marital status, a new graph has to be built from scratch. If, on the other hand, the cost function stays fixed and only new constraints (inconsistent with the current graph) are introduced, e.g., the counterfactuals should not be conditioned on a job change, the existing graph can be modified by removing some of its edges.} The Shortest Path First Algorithm (Dijkstra's algorithm) \cite{cormen2009introduction} is executed on the resulting graph over all the candidate targets, i.e., the set $\boldsymbol{I}_{CT}$ of all the data points that meet the confidence and density requirements (see line 11 in Algorithm~\ref{algo1}).

\newcommand\mycommfont[1]{\footnotesize\ttfamily\textcolor{black}{#1}}
\SetCommentSty{mycommfont}
\begin{algorithm}[ht]
\caption{\textbf{FACE} Counterfactual Generator}
\label{algo1}
\SetKwData{Left}{left}
\SetKwData{This}{this}
\SetKwData{Up}{up}
\SetKwFunction{Union}{Union}
\SetKwFunction{FindCompress}{FindCompress}
\SetKwInOut{Input}{input}
\SetKwInOut{Output}{output}
  \Input{Data ($\boldsymbol{X} \in \mathbb{R}^d)$, density estimator ($\hat{p}: \mathcal{X}\rightarrow[0, 1]$), probabilistic predictor ($\boldsymbol{clf}: \mathcal{X}\rightarrow[0, 1]$), distance function ($d: \mathcal{X}\times\mathcal{X}\rightarrow \mathbb{R}_{\scaleto{/geq0}{3pt}}$), distance threshold ($\epsilon > 0$), weight function ($w: \mathcal{X}\times\mathcal{X}\rightarrow \mathbb{R}_{\scaleto{>=0}{3pt}}$), and conditions function ($c: \mathcal{X}\times\mathcal{X}\rightarrow \{True,False\}$).}
\Output{Graph ($V, E, W$) and candidate targets ($\boldsymbol{I}_{CT}$).}

\BlankLine
\tcc{Construct a graph.}
\For{every pair ($\boldsymbol{x}_i, \boldsymbol{x}_j$) in $\boldsymbol{X}$}{
  \If{$d(\boldsymbol{x}_i, \boldsymbol{x}_j)$ $>$ $\epsilon$ $\boldsymbol{~and~}$ $c(\boldsymbol{x}_i, \boldsymbol{x}_j)$ $\boldsymbol{is}$ $\boldsymbol{\textit{True}}$}
    {
        $i \nsim j$\\
        $w_{ij} = 0$
    }
    \Else
    {
    	$i \sim j$ \\
        \tcc{In this case we use Equation~\ref{eq_kde} (KDE). This should be adjusted for $k$-NN and $\epsilon$-graph constructions by using Equation~\ref{eq_knn} and \ref{eq_egraph} respectively.}
        $w_{ij} = w(\hat{p}(\frac{\boldsymbol{x}_i + \boldsymbol{x}_j}{2})) \cdot d(\boldsymbol{x}_i, \boldsymbol{x}_j)$\\
    }
    }
    
\tcc{Get a set of candidate targets.}
$\boldsymbol{I}_{CT}$ = \{\} \\
\For{$\boldsymbol{x}_i$ in $\boldsymbol{X}$}{
\If{$\boldsymbol{clf}(\boldsymbol{x}_i) \geq t_p$ $\boldsymbol{~and~}$ $\hat{p}(\boldsymbol{x}_i) \geq t_d$}
    {
    $\boldsymbol{I}_{CT}$ = $\boldsymbol{I}_{CT}$ $\cup$ $i$
    }
}
\end{algorithm}

\paragraph{Complexity}
Execution of the Shortest Path First Algorithm between two instances can be optimised to have the worst case time complexity of $\mathcal{O}(|E| + |V|log|V|)$ where $|E|$ denotes the number of edges and $|V|$ the number of nodes in the graph. This complexity then scales accordingly to the number of candidate targets. The first term of the complexity -- the number of edges -- can be controlled by the user to a certain extent as it depends on the choice of the distance threshold parameter. The second term can also be controlled (and subsequently the first term as well) by reducing the number of instances to be considered, in which case the objective would be similar to the one of ``Prototype Selection''. A sub-sampling as simple as a random sampling of the data points, or more sophisticated alternatives such as Maximum Mean Discrepancy \cite{kim2016examples,gretton2012kernel}, can be used with a clear trade-off between the accuracy of the generated counterfactuals and the algorithm's speed.%

In practice a base graph can be generated and stored with the most generic conditions imposed, e.g., if the data represent people, edges between people of different sex would be removed. When an explainee requests a counterfactual, he can impose further restrictions (by removing edges) to create a personalised graph, e.g., this individual is not willing to get divorced. On the other hand, if personalised cost function is required, entirely new graph needs to be generated. While the theory presented here only holds for continuous distributions, which satisfy the requirements discussed earlier, the approach can still be used with discrete features.

\section{Experiments\label{experiments}}
To empirically demonstrate the utility of \textbf{FACE} we present results of its execution on two distinct data sets. First, we show the behaviour of our algorithm on a toy data set and compare the three graph construction approaches introduced in Section~\ref{methods}. Secondly, we apply our algorithm to the MNIST data set \cite{lecun2010mnist} and show how it can be used to derive meaningful digit transformations based on the calculated path.

\begin{figure}
\centering
\begin{subfigure}[t]{.47\textwidth}
  \centering
  \includegraphics[width=1.\linewidth]{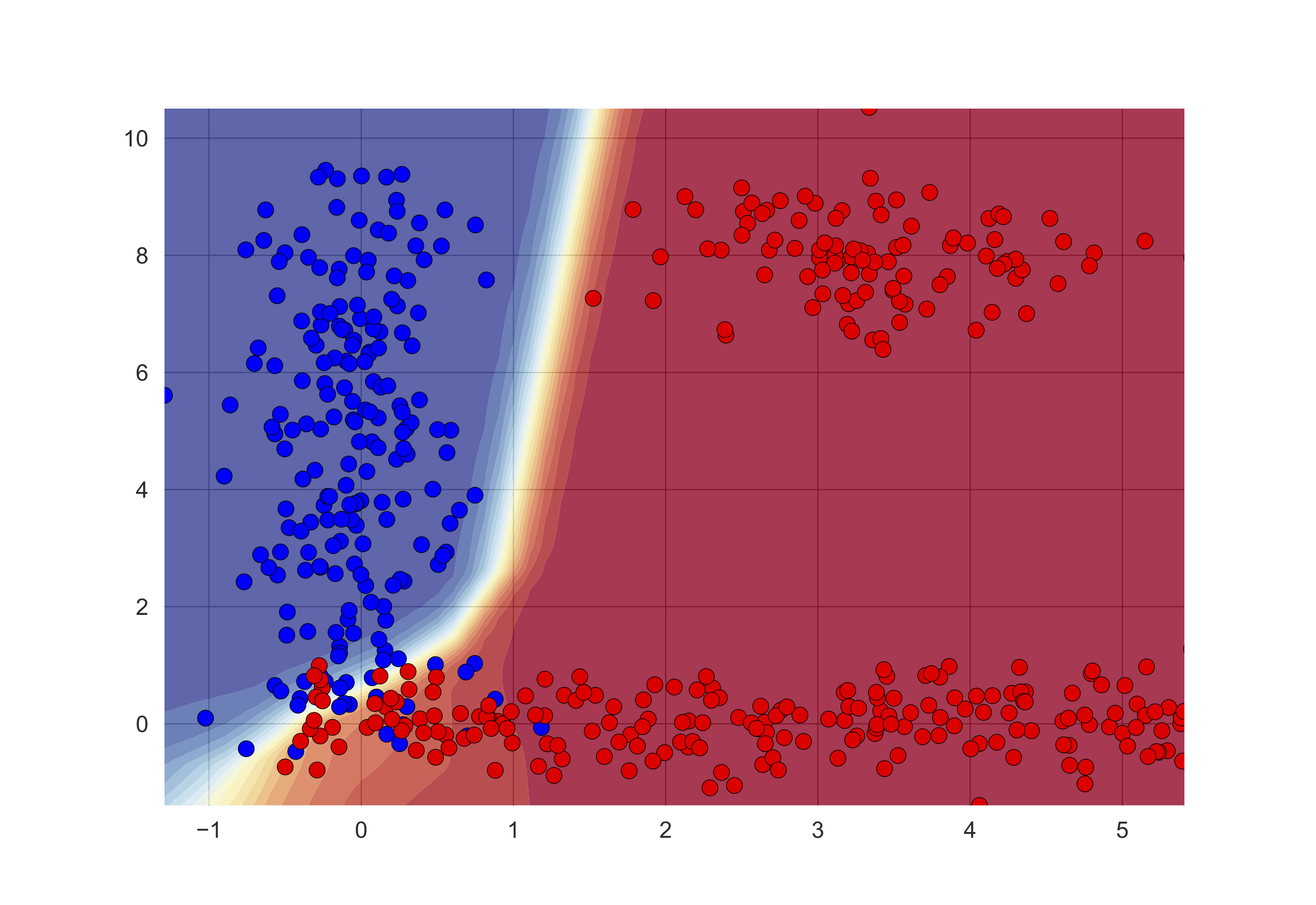}
  \caption{$\epsilon = 0.25$ distance threshold.}
\end{subfigure}%
\newline
\begin{subfigure}[t]{.47\textwidth}
  \centering
  \includegraphics[width=1.\linewidth]{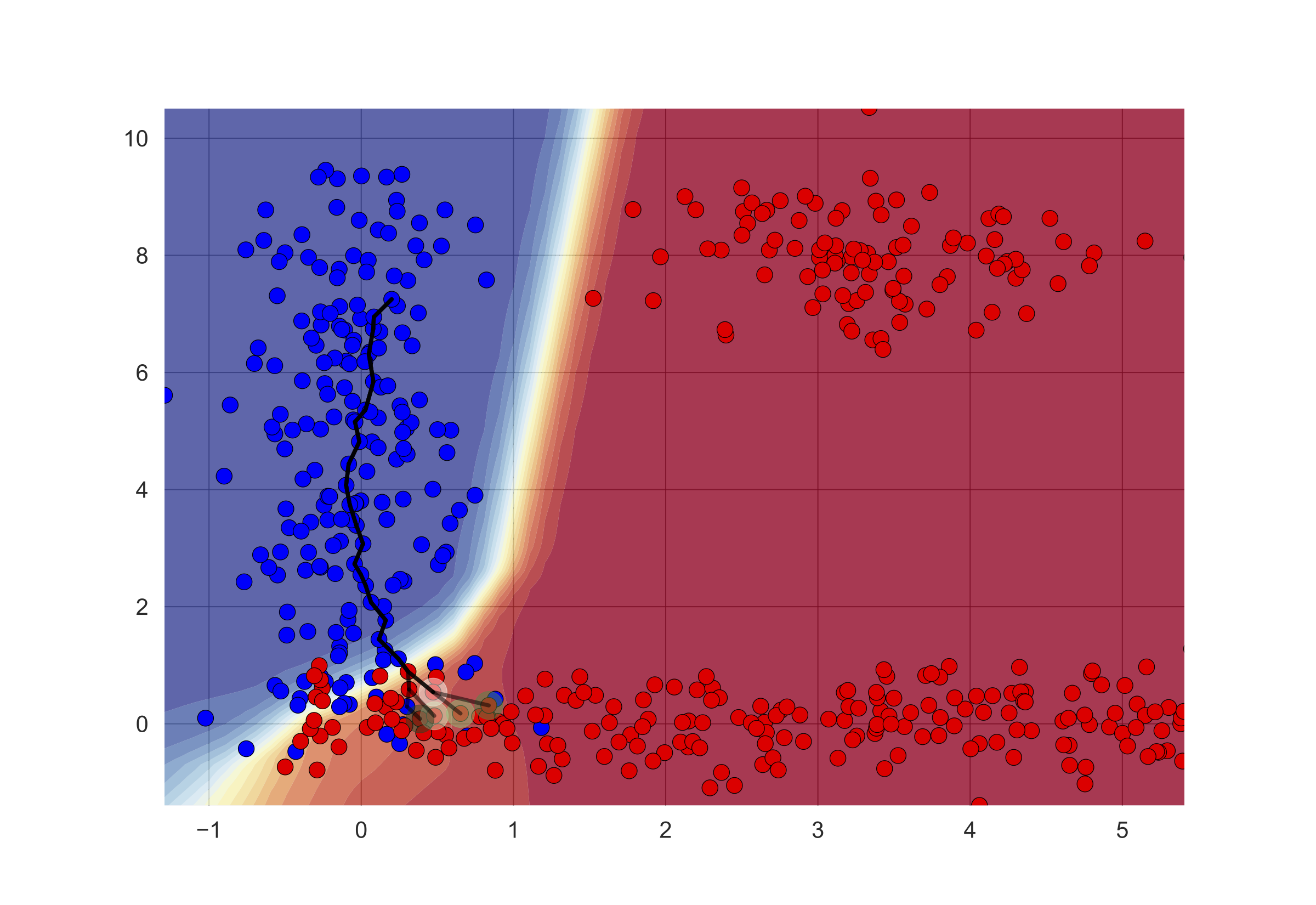}
  \caption{$\epsilon = 0.50$ distance threshold.}
\end{subfigure}%
\newline
\begin{subfigure}[t]{.47\textwidth}
  \centering
  \includegraphics[width=1.\linewidth]{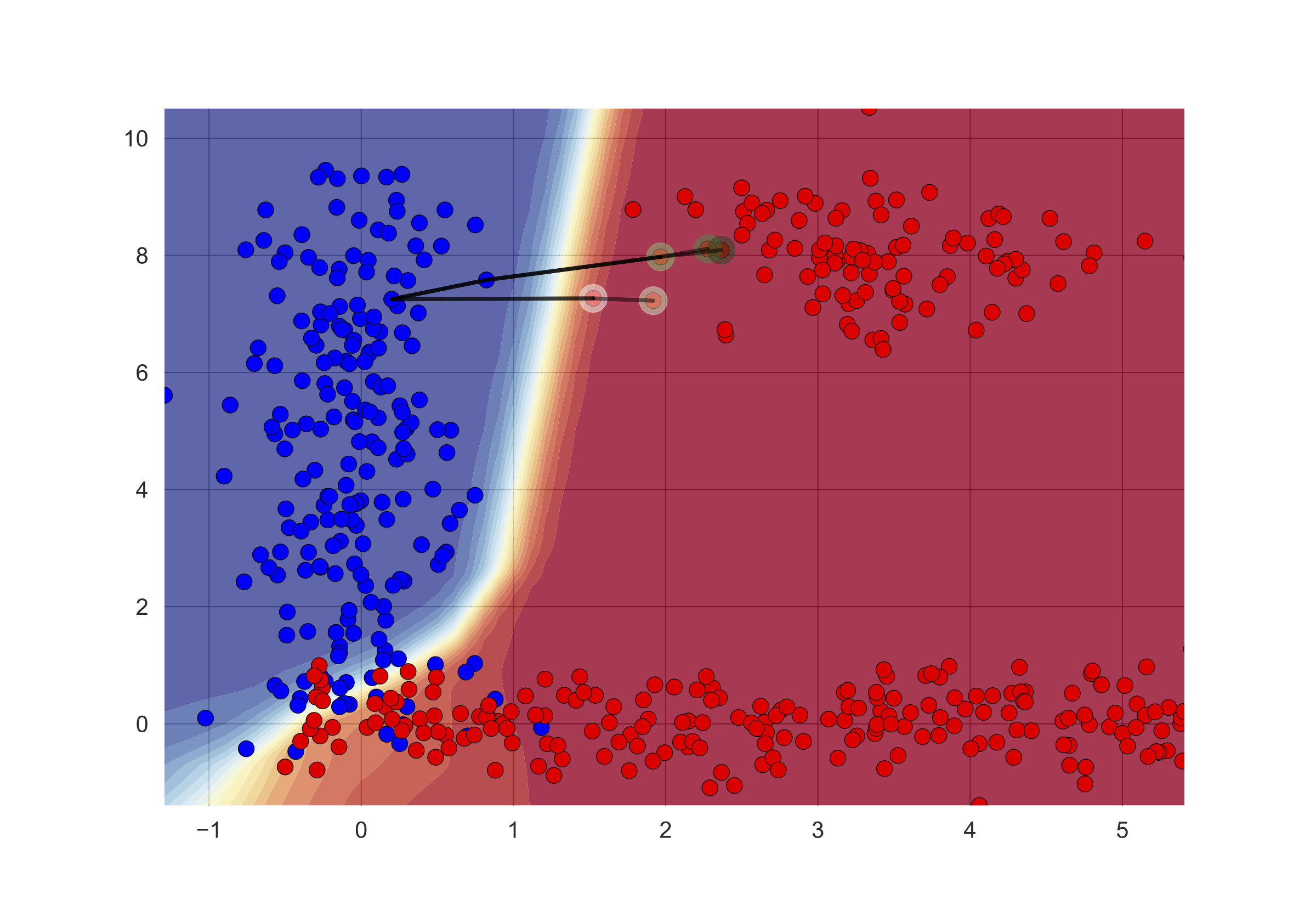}
  \caption{$\epsilon = 2$ distance threshold.}
\end{subfigure}%
\caption{The five shortest paths from a starting data point to a target (counterfactual) data point generated from a graph, which edge weights were computed using the $KDE$ approach. The targets are restricted by: i) $t_p \geq 0.75$ prediction threshold, ii) $t_d \geq 0.001$ density threshold.}
\label{fig:kde_2d}
\end{figure}

\begin{figure}
\centering
\begin{subfigure}[t]{.47\textwidth}
  \centering
  \includegraphics[width=1.\linewidth]{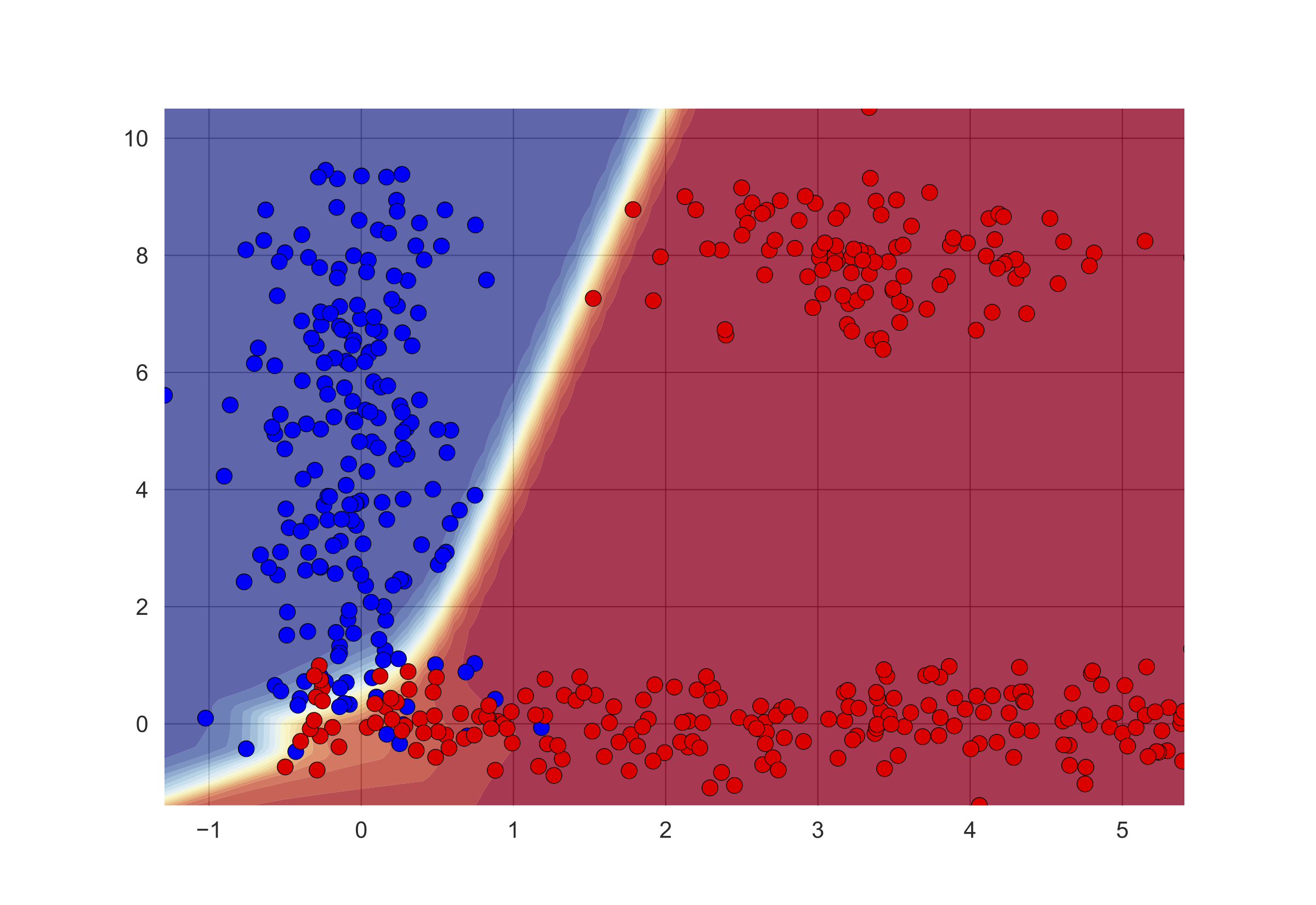}
  \caption{$\epsilon = 0.25$ distance threshold.}
\end{subfigure}%
\newline
\begin{subfigure}[t]{.47\textwidth}
  \centering
  \includegraphics[width=1.\linewidth]{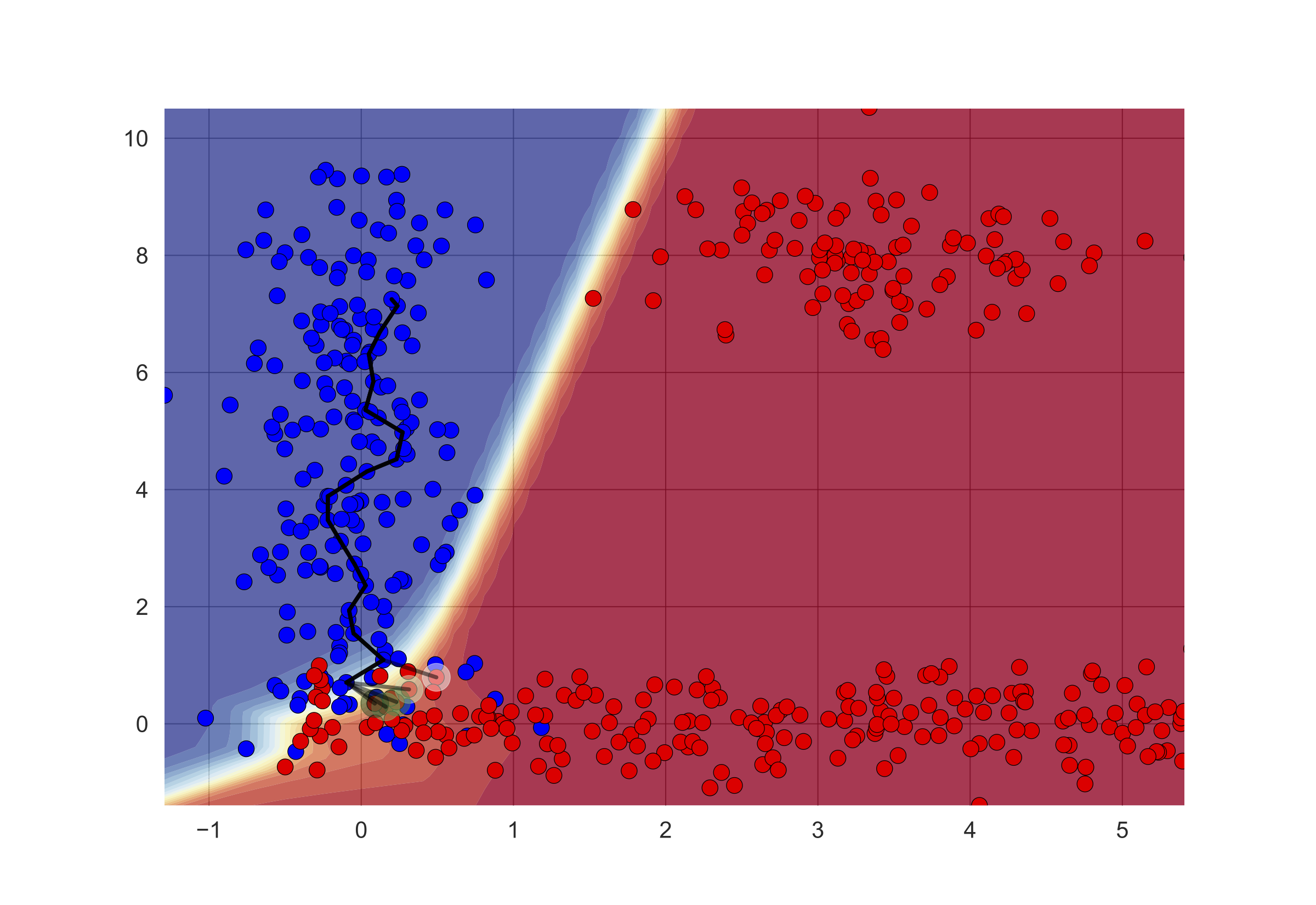}
  \caption{$\epsilon = 0.50$ distance threshold.}
\end{subfigure}
\newline
\begin{subfigure}[t]{.47\textwidth}
  \centering
  \includegraphics[width=1.\linewidth]{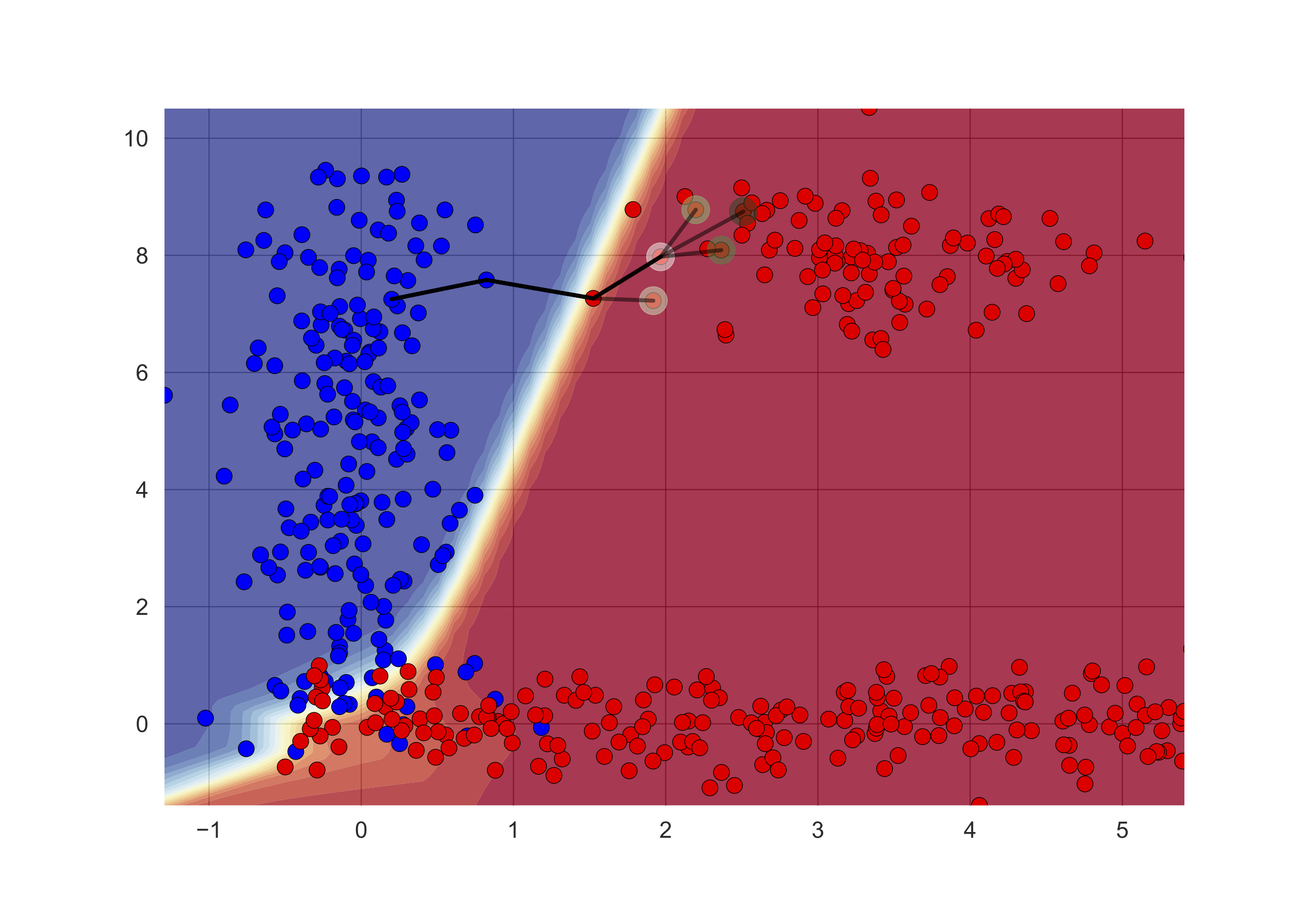}
  \caption{$\epsilon = 1$ distance threshold.}
\end{subfigure}
\caption{The five shortest paths from a starting data point to a target (counterfactual) data point generated from a graph, which edge weights were computed using the $e$-graph approach. The targets are restricted by $t_p \geq 0.75$ prediction threshold.}
\label{fig:egraph}
\end{figure}

\begin{figure}
\centering
\begin{subfigure}{.47\textwidth}
  \centering
  \includegraphics[width=1.\linewidth]{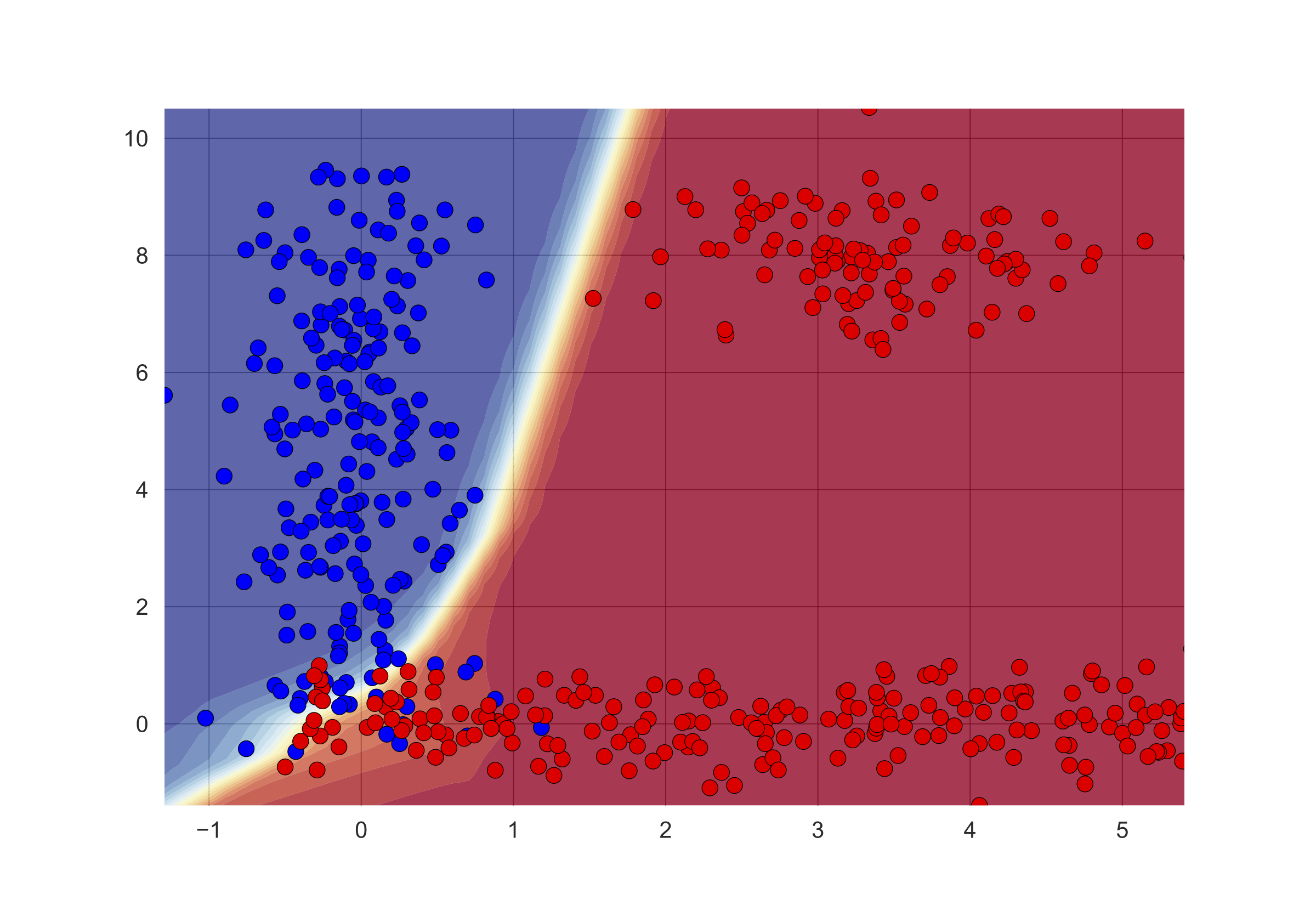}
  \caption{$k = 2$ neighbours and $\epsilon = 0.25$ distance threshold.}
\end{subfigure}%
\newline
\begin{subfigure}{.47\textwidth}
  \centering
  \includegraphics[width=1.\linewidth]{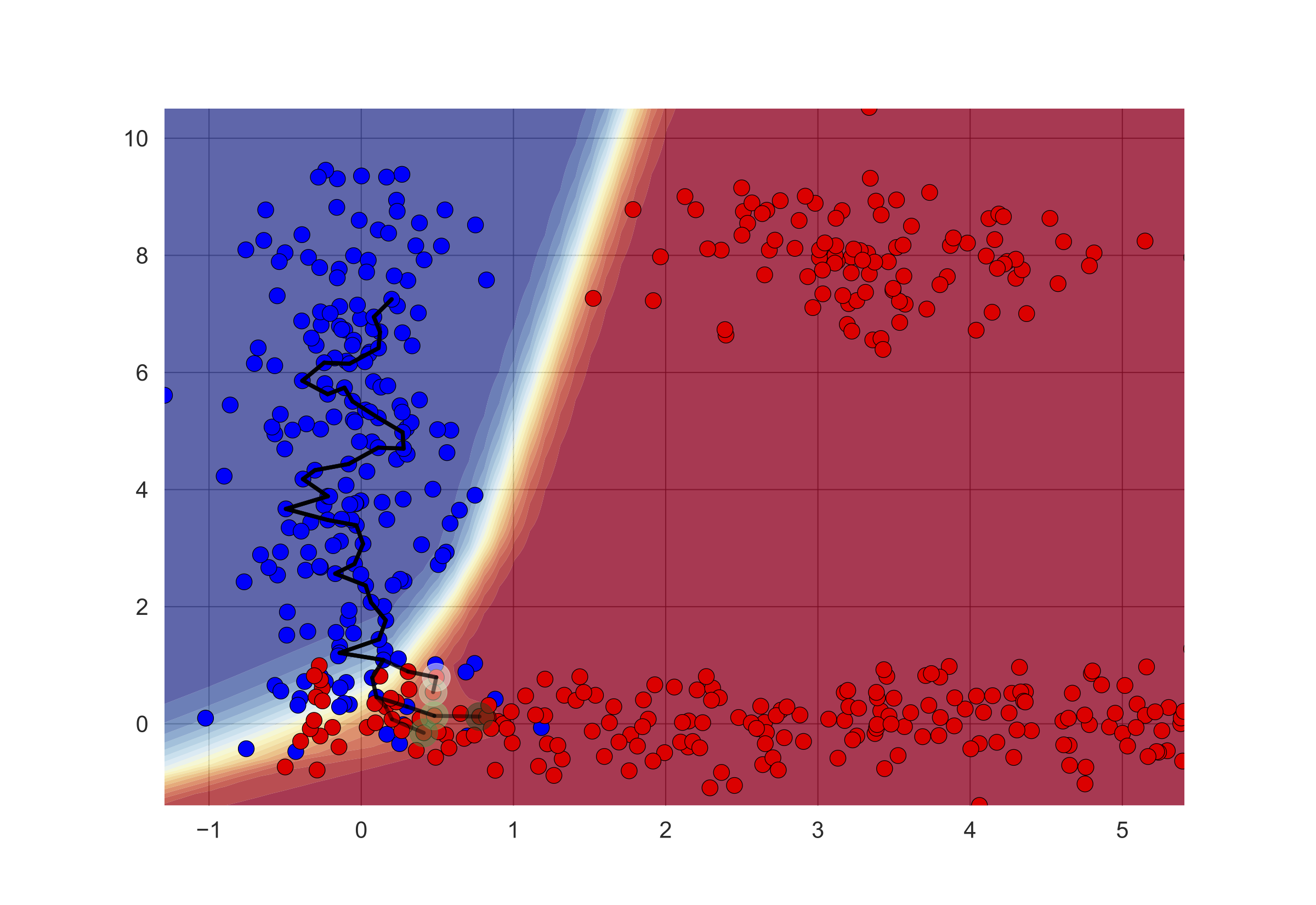}
  \caption{$k = 4$ neighbours and $\epsilon = 0.35$ distance threshold.}
\end{subfigure}
\newline
\begin{subfigure}{.47\textwidth}
  \centering
  \includegraphics[width=1.\linewidth]{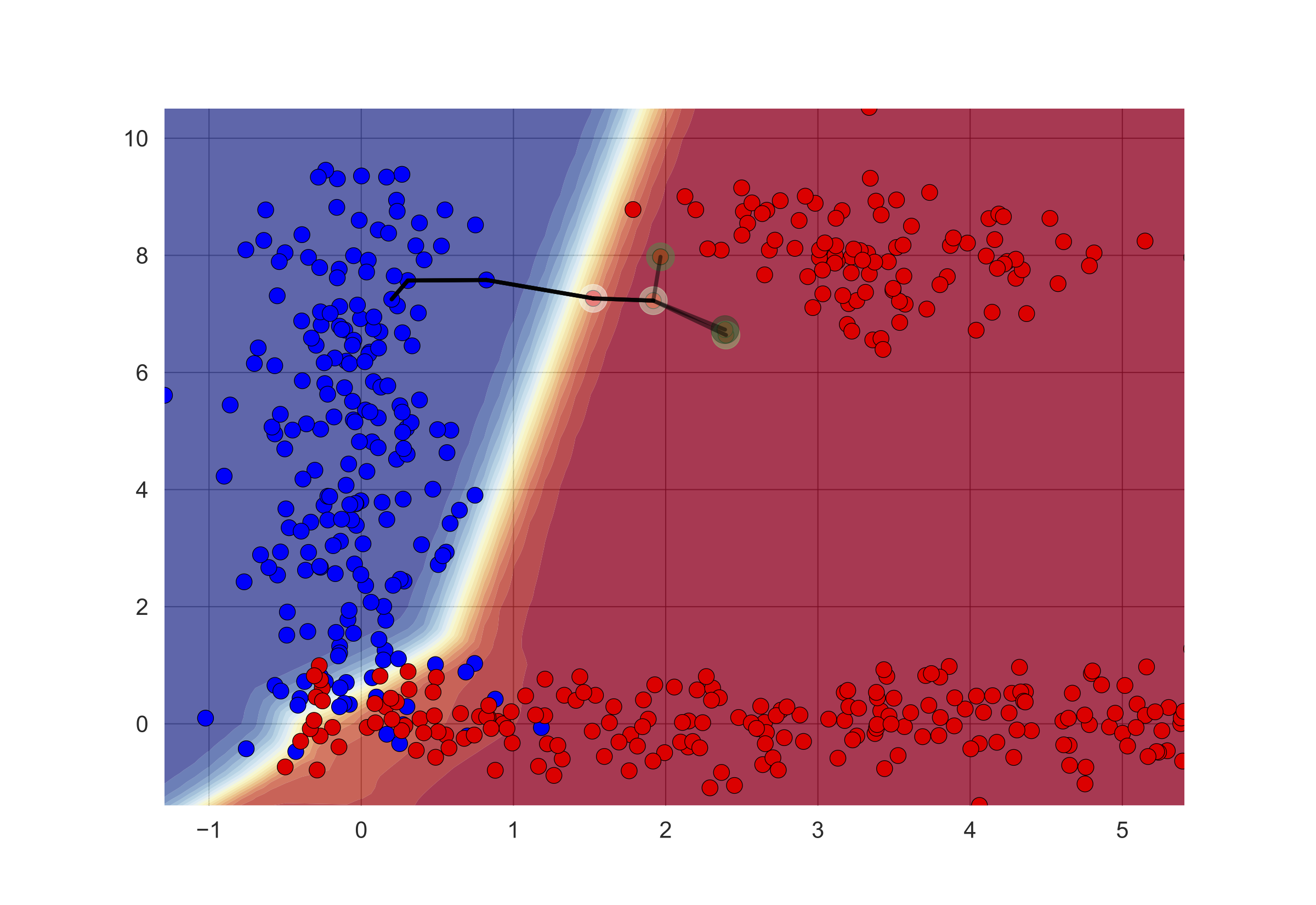}
  \caption{$k = 10$ neighbours and $\epsilon = 0.80$ distance threshold.}
\end{subfigure}
\caption{The five shortest paths from a starting data point to a target (counterfactual) data point generated from a graph, which edge weights were computed using the $k$-NN graph approach. The targets are restricted by $t_p \geq 0.75$ prediction threshold with the $\epsilon$ distance threshold and $k$ neighbours set to: (a) $k = 2$ and $\epsilon = 0.25$; (b) $k = 4$.}
\label{fig:knn}
\end{figure}

\begin{figure}
  \centering
  \includegraphics[width=1.\linewidth]{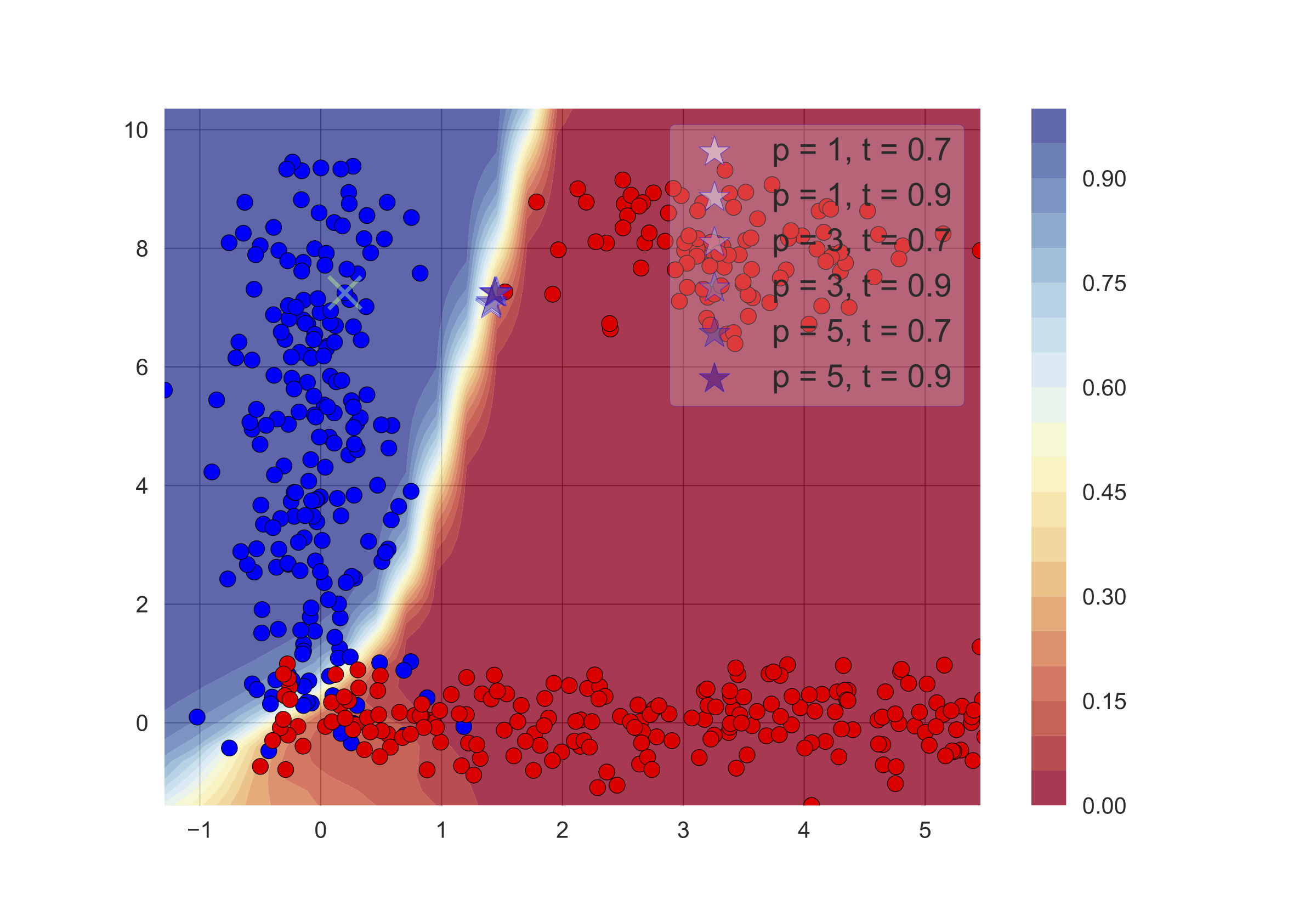}
  \caption{Counterfactuals generated using the method proposed by \citet{wachter2017counterfactual}. $p$ denotes the penalty parameter and $t$ the classification threshold. These counterfactuals clearly do not comply with the desired properties described in Section~\ref{counterfactuals}.}
  \label{fig:oxford_paper}
\end{figure}

\paragraph{Synthetic Data Set}
To this end, we trained a Neural Network with two hidden layers of length 10 and ReLU activation functions. %
%
\textbf{FACE} was initialised with $w(z)=-log(z)$ as the weight function and the $l_2$-norm as the distance function. Figures~\ref{fig:kde_2d}, \ref{fig:egraph} and \ref{fig:knn} show the results of applying \textbf{FACE} to the toy data set when used with $KDE$, $e$-graph and $k$-NN respectively.
In each, the triplet follows a similar pattern: (a) no counterfactual is generated, (b) a ``good'' counterfactual is generated, and (c) a ``bad'' counterfactual is generated. Our experimental setup adheres to a real-life use case where \textbf{FACE} is originally applied with a fairly ``restrictive'' configuration, which is subsequently being relaxed until a counterfactual is found. Figure~\ref{fig:oxford_paper} shows the counterfactuals found by optimising Equation~\ref{adv_examples} proposed by \citet{wachter2017counterfactual}, which can be compared against the ones achieved with \textbf{FACE} on the same data set (cf.\ Figures~\ref{fig:kde_2d}, \ref{fig:egraph} and \ref{fig:knn}).

%

\begin{figure}[t]
\centering
\begin{subfigure}[b]{0.45\textwidth}
   \includegraphics[width=1\linewidth]{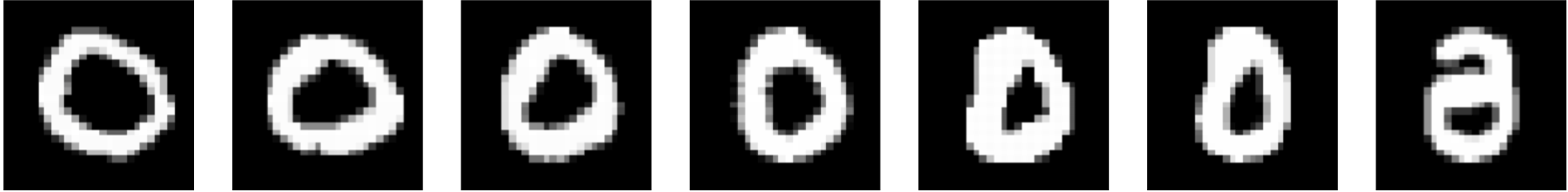}
   \label{fig:Ng1} 
\end{subfigure}

\begin{subfigure}[b]{0.45\textwidth}
   \includegraphics[width=1\linewidth]{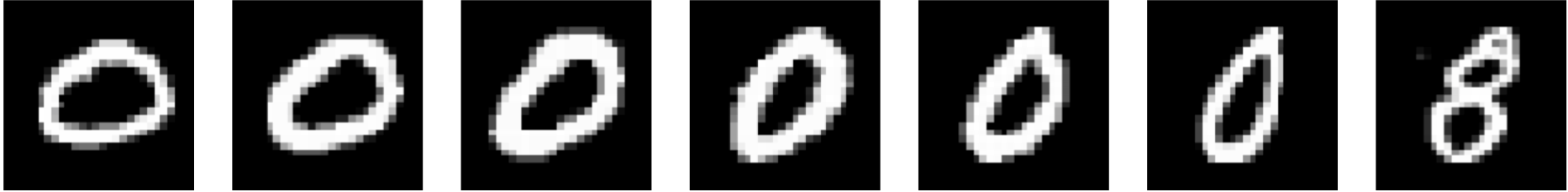}
   \label{fig:Ng2}
\end{subfigure}

\caption{The \textbf{FACE} ``transformation'' paths between zero and eight (counterfactual) for two different MNIST images.}
\label{mnist}
\end{figure}

\paragraph{MNIST Data Set}
We applied \textbf{FACE} (based on the $k$-NN construction algorithm with $k=50$) to two images of the zero digit from the MNIST data set \cite{lecun2010mnist} with the target counterfactual class set to the digit eight.
The underlying predictive model is a Neural Network. Figure~\ref{mnist} depicts the full path from the starting instance (left) to the final counterfactual (right). The resulting path shows a smooth transformation through the zeros until an eight is reached.

\section{Related Work\label{related_work}}
Counterfactual Explanations have been deemed to satisfy the ``Right to Explanation'' requirement \cite{wachter2017counterfactual} introduced by the European Union's General Data Protection Regulation (GDPR), making them viable for many businesses applying predictive modelling to human matters. %
To this end, \citet{wachter2017counterfactual} adapted machinery used in the \textit{Adversarial Examples} literature \cite{goodfellow2015explaining}:
\begin{equation}
    \label{adv_examples}
    \arg \min_{x'} \max_{\lambda}  (f_w(\boldsymbol{x'}) - y')^2 + \lambda \cdot d(\boldsymbol{x}, \boldsymbol{x'}) \text{,}
\end{equation}
where $\boldsymbol{x}$ and $\boldsymbol{x'}$ denote respectively the current state of the subject and the counterfactual, $y'$ the desired outcome, $d(\cdot, \cdot)$ a distance function and $f_w$ a classifier parametrised by $w$. The objective is optimised by iteratively solving for $\boldsymbol{x'}$ and increasing $\lambda$ until a sufficient solution is found. %
\citeauthor{wachter2017counterfactual} emphasise the importance of the distance function choice and suggest using the $l_1$-norm penalty on the counterfactual, to induce sparse solutions, weighted by the \textit{Median Absolute Deviation}. The authors deal with discrete variables by doing a separate execution of the optimisation problem, one for each unique value of every feature, and then choosing a counterfactual with the shortest distance.%


\citet{ustun2018actionable} present an Integer Programming toolkit for linear models that can be used by practitioners to analyse actionability and difficulty of recourse in a given population as well as generate advice for actionable changes (counterfactuals). Their tool ``ensures recourse [actionability] in linear classification problems without interfering in model development''~\cite{ustun2018actionable} but it does not take into account: (1) counterfactuals residing in high-density regions and (2) the existence of high-density paths connecting explained data points with counterfactual examples.%

\citet{efficient2019russel} propose a Mixed Integer Programming (MIP) formulation to handle mixed data types and offer counterfactual explanations for linear classifiers that respect the original data structure. This formulation is guaranteed to find coherent solutions (avoiding nonsense states) by only searching within the ``mixed-polytope'' structure defined by a suitable choice of linear constraints. \citet{efficient2019russel} chose an iterative approach to providing diverse collection of counterfactuals. Given one solution, the user can add extra constraints to the MIP that will restrict previous alterations. The list of counterfactuals is then ranked according to their $l_1$-distance to the explained instance.%

\citet{van2018contrastive} propose a counterfactual generation method for decision trees. Their approach uses locally trained one-vs-rest decision trees to establish a set of disjoint rules that cause the chosen instance to be classified as the target class.

\textbf{FACE} improves over all of the aforementioned counterfactual generation schemata in a number of ways:
\begin{itemize}
    \item contrarily to \citet{wachter2017counterfactual} and similarly to \citet{ustun2018actionable}, \citet{efficient2019russel} and \citet{van2018contrastive} it supports discrete features and their restrictions in a principled manner;
    \item contrarily to \citet{ustun2018actionable}, \citet{efficient2019russel} and \citet{van2018contrastive}, and similarly to \citet{wachter2017counterfactual} it is \emph{model-agnostic}; and
    \item contrarily to all four approaches it produces counterfactual explanations that are both feasible and actionable.
\end{itemize}
\section{Summary and Future Work\label{discussion}}
In this paper we have highlighted the shortcomings of popular Counterfactual Explanation approaches in the Machine Learning literature and proposed a new method, called \textbf{FACE}, that aims at resolving them. Our approach accounts for both the nature of the the counterfactual and the degree to which the proposed change is feasible and actionable. %
Our future work includes the performance evaluation of \textbf{FACE} %
on real-world data sets of dynamic nature and exploring the degree to which our suggested counterfactuals match the \textit{true} change.%
%

\begin{acks}
This research was supported by the Flemish Government under the ``Onderzoeksprogramma Artifici\"ele Intelligentie (AI) Vlaanderen'' programme, the EPSRC (SPHERE EP/R005273/1) and the MRC Momentum award (CUBOID MC/PC/16029).
\end{acks}

\bibliographystyle{ACM-Reference-Format}
\balance 
\bibliography{bibliography}

\end{document}